\documentclass[10pt,twocolumn,letterpaper]{article}

\usepackage[pagenumbers]{wacv} 

\usepackage[table]{xcolor}
\definecolor{softred}{RGB}{252,235,235}
\definecolor{softyellow}{RGB}{255,249,225}
\definecolor{softgreen}{RGB}{235,247,235}
\newcommand{\cbox}[1]{%
  \fcolorbox{black}{#1}{\rule{0pt}{0.4em}\rule{0.4em}{0pt}}
}

\definecolor{wacvblue}{rgb}{0.21,0.49,0.74}
\usepackage{comment}
\usepackage{algorithm}
\usepackage{algpseudocode}
\usepackage{multirow}
\usepackage{adjustbox} 
\usepackage{soul}
\usepackage[pagebackref,breaklinks,colorlinks,allcolors=wacvblue]{hyperref}

\usepackage[table]{xcolor}
\definecolor{softred}{RGB}{252,235,235}
\definecolor{softyellow}{RGB}{255,249,225}
\definecolor{softgreen}{RGB}{235,247,235}

\def\wacvPaperID{1367} 
\def\confName{WACV}
\def\confYear{2027}

\title{Benchmarking Nighttime Traffic Sign Recognition with Illumination-Adaptive Detection and Semantic Attribute Reasoning}

\author{Aditya Mishra\\
IISER Bhopal\\
{\tt\small aditya21@iiserb.ac.in}
\and
Haroon R. Lone\\
IISER Bhopal\\
{\tt\small haroon@iiserb.ac.in}
}

\begin{document}
\maketitle
\begin{abstract}
Traffic signboards are vital for road safety and intelligent transportation systems. Yet, recognizing traffic signs at night remains underexplored due to the scarcity of real-world public datasets capturing low-light degradations and distractor classes. Existing benchmarks are predominantly daytime and do not reflect challenges such as headlight glare, motion blur, sensor noise, and vandalized or ambiguous signage. To address these gaps, we introduce INTSD, a large-scale nighttime traffic sign dataset collected across diverse regions of India. INTSD contains street-level images spanning 41 traffic signboard classes, multiple distractor categories, and varied lighting and weather conditions, designed to support both detection and fine-grained classification under nighttime scenarios. To benchmark INTSD, we conduct extensive evaluations using state-of-the-art detection and classification models under standardized protocols. Additionally, we present LENS-Net, a strong baseline that integrates an end-to-end adaptive illumination-aware detector with a multimodal classifier that fuses vision-language representations with soft semantic attribute reasoning over learnable shape and color embeddings. Experiments demonstrate that models trained exclusively on daytime data fail substantially under real nighttime conditions — a gap that is recovered once INTSD is introduced in training, even when controlling for data volume. These results validate INTSD as a complementary nighttime training resource and establish competitive baselines for future research.
The code and dataset are publicly available\footnote{\url{https://adityamishra-ml.github.io/INTSD/}}.
\end{abstract}
    
\section{Introduction}
\label{sec:intro}


\begin{table*}[t]
\centering
\normalsize
\caption{Overview of traffic sign datasets. $^{\ddag}$ TT100K has only 10000 images with signboard in it. $^{\S}$video-frames covering only 15,630 unique signs. $^{\P}$45 classes have more than 100 examples. $^{\intercal}$ The MVD data set only contains front and back signboard classes. \textit{Unique} means that the same physical signboard does not appear repeatedly in multiple images. \textit{Nighttime} denotes datasets primarily collected under nighttime conditions.} 
\label{tab:data_comparision}
\setlength{\tabcolsep}{4pt}
\resizebox{1\textwidth}{!}{
\begin{tabular}{lcccp{3cm}p{2cm}ccc}
\hline
\textbf{Dataset} & \textbf{Images} & \textbf{Classes} & \textbf{Signs} & \textbf{Attributes} & \textbf{Region} & \textbf{Boxes} & \textbf{Unique} & \textbf{Nighttime}\\
\hline
\textbf{INTSD} & 11016 & 41 & 14,044 & tampered, blur, glare,  & India & $\checkmark$ & $\times$ & $\checkmark$\\

 &  &  &  & advertisements, &  &  &  &  \\

&  &  &  & MTSD's attributes &  &  &  & \\
\hline
CNTSSS \cite{lin2025yolo}  & 4062 & 3 & 7687 & \centering{$\times$}  & China & $\checkmark$ & $\times$ & $\checkmark$ \\
\hline
MTSD \cite{ertler2020mapillary} & 105,830 & 400 & 354,154 & occluded, exterior, dummy, \hspace{-0.98mm}out-of-frame, & Global & $\checkmark$ & $\checkmark$ & $\times$ \\
  &  &  &  & included, ambiguous &  &  &  \\
TT100K \cite{zhu2016traffic}  & $^{\ddag}$100,000 & $^{\P}$221 & 26,349 & \centering{$\times$} & China & $\checkmark$ & $\checkmark$ & $\times$ \\
MVD \cite{neuhold2017mapillary}  & 20,000 & $^{\intercal}$2 & 174,541 & \centering{$\times$} & Global & $\checkmark$ & $\checkmark$ & $\times$ \\
BDD100K \cite{yu2018bdd100k}  & 100,000 & 1 & 343,777 & \centering{$\times$} & USA & $\checkmark$ & $\times$
& $\times$ \\
\hline
GTSDB \cite{stallkamp2011german} & 900 & 43 & 852 & \centering{$\times$} & Germany & $\checkmark$  & $\times$ & $\times$ \\
RTSD \cite{shakhuro2016russian}  & $^{\S}$179,138 & 156 & $^{\S}$104,358 & \centering{$\times$} & Russia & $\checkmark$ & $\times$ & $\times$ \\
STS \cite{larsson2011using}  & 3777 & 20 & 5582 & \centering{$\times$} & Sweden & $\checkmark$ & $\times$ & $\times$ \\
LISA \cite{mogelmose2012vision}  & 6610 & 47 & 7855 & \centering{$\times$} & USA & $\checkmark$ & $\times$ & $\times$  \\
GTSRB \cite{stallkamp2012man}  & $\times$ & 43 & 39,210 & \centering{$\times$} & Germany & $\times$ & $\checkmark$ & $\times$  \\
BelgiumTS \cite{mathias2013traffic} & $\times$ & 108 & 8851 & \centering{$\times$} & Belgium & $\times$ & $\times$ & $\times$  \\
\hline
\end{tabular}
}
\end{table*}

Traffic signboards play a critical role in Advanced Driver Assistance Systems (ADAS), autonomous driving, navigation, and smart city systems. However, most existing traffic sign datasets \cite{shakhuro2016russian, zhu2016traffic, ertler2020mapillary} predominantly feature daytime imagery, leaving nighttime recognition underexplored. To mitigate this gap, prior studies have used generative translation or domain adaptation techniques such as CycleGAN \cite{zhu2017unpaired} and CyCADA \cite{hoffman2018cycada} to synthesize nighttime scenes from daytime data. Yet, these approaches often introduce unrealistic artifacts and fail to capture complex photometric effects (such as glare, reflections, and sensor noise), resulting in a persistent domain gap that hinders robustness in real-world deployment \cite{sakaridis2021acdc, chen2018learning, punnappurath2022day}. Additionally, a prior study shows that generative translation to nighttime conditions do not accurately assess a model’s performance during the real nighttime scenario \cite{lin2025yolo}. 

TSR has been widely studied in computer vision research \cite{li2017perceptual, lin2025yolo, eykholt2018robust, sun2024llth}. However, these studies have largely been validated on datasets with significant limitations. These benchmarks typically lack nighttime imagery, data from diverse camera sensors, and varied weather conditions. Moreover, the traffic sign classes represented often exclude real-world adversaries, such as damaged or partially occluded signs. This presents a gap, underscoring the need for extensive research under more challenging and realistic operational scenarios, particularly during nighttime. 

To address this limitation, we present INTSD, a real-world \textbf{I}ndian \textbf{N}ighttime \textbf{T}raffic \textbf{S}ign \textbf{D}ataset. The INTSD covers a wide range of geographical area across India and features 14k+ fully annotated instances, in diverse lighting and weather conditions. It also introduces four additional classes that are frequently misclassified by TSR models but are absent in existing benchmarks. The inclusion of natural noise factors such as glare and motion blur, makes it a new standard for traffic sign detection and classification tasks. 

A central question for any new dataset is whether it provides information that existing data does not. We answer this directly: models trained only on daytime Indian traffic signs degrade sharply when evaluated on real nighttime scenes, and this gap is recovered once INTSD is introduced into training. Across three architecturally distinct recognizers, daytime-only training yields weak nighttime performance: detection collapses almost entirely (e.g., mAP@50 of 5.31\% for YOLO-TS), and classification degrades sharply (54--71\% macro-precision), whereas adding INTSD restores both to the 90\%+ range. Crucially, this recovery persists even when the injected nighttime data is subsampled to match the daytime set in size, isolating the effect of the nighttime \emph{signal} from that of additional data volume. These results establish that the low-light degradations captured by INTSD---glare, motion blur, sensor noise, and underexposure---are not recoverable from daytime imagery alone, and that INTSD supplies training signal complementary to existing daytime benchmarks.

To accompany the benchmark, we additionally provide LENS-Net (\textbf{L}earned \textbf{E}nhancement \& \textbf{N}eural \textbf{S}emantics \textbf{Net}work), a reference baseline that follows a detect-then-classify design: (i) a detector that couples an end-to-end, per-image adaptive illumination-enhancement module with localization, and (ii) a classifier that fuses frozen vision-language (CLIP) representations with soft semantic attribute reasoning over learnable shape and color embeddings. We present LENS-Net not as the central contribution but as a competitive, illumination-aware baseline that probes which design choices matter under nighttime degradation; our ablations indicate that the gains stem from semantic vision-language priors rather than from architectural complexity. Alongside LENS-Net, we benchmark a broad set of standard detectors and classifiers under a single standardized protocol, so that INTSD can serve as a fair and reproducible evaluation platform.

In summary, we make the following contributions:

\begin{enumerate}

    \item We introduce INTSD, the first large-scale, diverse Indian nighttime traffic sign dataset: 11k+ images and 14k+ fully annotated instances across 41 traffic-sign and distractor classes, captured under varied sensors, weather, and illumination, with attribute-level annotations. 

    \item We establish a standardized benchmark on INTSD for both detection and classification, evaluating a broad range of contemporary models under a unified protocol, together with a cross-domain utility analysis demonstrating that daytime-trained models fail at night and that INTSD recovers their performance.

    \item We provide LENS-Net, an illumination-aware multimodal reference baseline, and use its ablations to identify which components contribute to robust recognition under low-light conditions.

\end{enumerate}

\section{Literature Review}
\label{sec:rel_work}

A wide range of traffic sign datasets support detection and recognition research. Table \ref{tab:data_comparision} provides overview and comparison of publicly available benchmarks. Early benchmarks such as GTSRB \cite{stallkamp2012man} and GTSDB \cite{stallkamp2011german} focused on well-lit, structured settings, while later datasets like SwedishTS \cite{larsson2011using}, BelgiumTS \cite{mathias2013traffic}, RTSD \cite{shakhuro2016russian}, and TT100K \cite{zhu2016traffic} expanded geographic and environmental diversity. Large-scale datasets including  MVD \cite{neuhold2017mapillary}, BDD100K \cite{yu2018bdd100k}, and MTSD \cite{ertler2020mapillary} introduced complex contextual challenges. Yet, nighttime datasets remain scarce, motivating INTSD, a dedicated benchmark for low-light traffic sign detection and recognition. CNTSSS \cite{lin2025yolo} recently addressed nighttime detection with a limited set of traffic sign categories. INTSD extends this direction by enabling both detection and fine-grained classification by incorporating realistic distractor categories encountered in real-world driving.

\noindent \textbf{Image Enhancement:} The nighttime conditions adds noise and taint the computer vision tasks. Prior studies \cite{Tran_2024_CVPR} have addressed this through image enhancement techniques aimed at improving visibility before detection. Recent works~\cite{liu2022image,cui2022you} proposed adaptive and restoration-based frameworks that jointly optimize enhancement and detection, while zero-reference mapping~\cite{guo2020zero} and illumination-aware methods~\cite{sun2024llth} also show notable gains in nighttime perception. Motivated by these advances, LENS-Net detector integrates adaptive image enhancement, enabling end-to-end illumination correction tuned for sign localization.

\noindent \textbf{Detection:} Detection has evolved from heuristic color-shape approaches \cite{han2017traffic, sun2019traffic} to deep CNN-based architectures. One-stage detectors \cite{lin2017focal, liu2016ssd, redmon2017yolo9000} overcome the slowness of two-stage detectors \cite{ren2015faster, dai2016r, lin2017feature, girshick2014rich}, however, cluttered nighttime signboards still challenge these models, motivating adaptive and illumination-aware detection.

\noindent \textbf{Classification:} Following detection, classification remains a fine-grained visual challenge, as many signs differ only in subtle symbols or numerals (e.g., no parking vs. parking). CNN-based classifiers have long dominated this space, outperforming earlier hand-crafted pipelines \cite{zhu2016traffic, haloi2015traffic}, while recent approaches use attention \cite{chang2023erudite, yang2018learning} and transformer-based \cite{farzipour2023traffic} architectures to focus on discriminative regions. Such mechanisms are critical for distinguishing visually similar signs, especially under low-light or degraded conditions. In this work, the proposed  CLIP-MLP based classifier builds upon this insight, combining cross-modal attention with attribute reasoning for semantically coherent classification.


\section{Dataset}
\label{sec:data}
\begin{figure}[t]
    \centering
    \includegraphics[width=0.9\columnwidth]{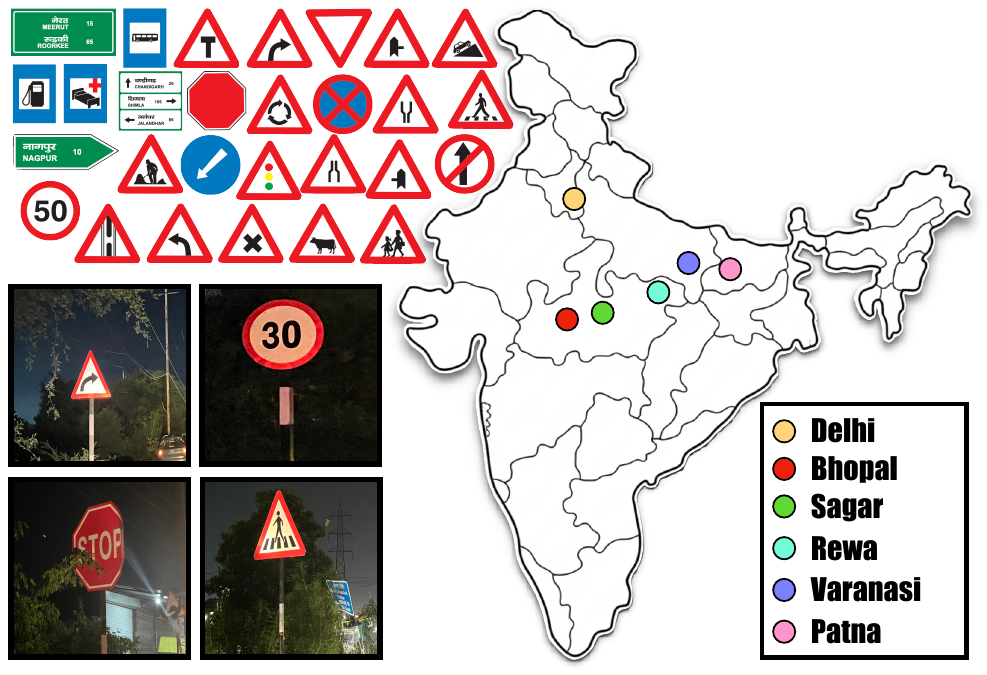}
    \caption{Geographical coverage and representative nighttime traffic sign examples from INTSD. The dataset spans six districts and includes both rural and urban traffic signs captured under diverse lighting and weather conditions.}
    \label{fig:data-1.pdf}
\end{figure}

Traffic sign distributions vary widely across regions due to differences in urban, highway, and rural infrastructure \cite{ertler2020mapillary}. To capture this diversity, the INTSD spans six districts across India, encompassing urban and rural settings and varying environmental conditions such as weather, illumination, motion blur, and glare. Each image contains multiple sign instances, ensuring dense spatial coverage. Figure~\ref{fig:data-1.pdf} illustrates representative examples and their geographic distribution. The INTSD also includes challenging scenarios in which signboards are partially occluded by foliage, obscured by other signs, partially out-of-frame, or form part of larger signboards.  

\subsection{Data Collection and Curation}
The images are primarily extracted from videos recorded at 30fps using a low-budget ($\approx 60$ USD) action camera (AKASO Ek7000), supplemented with images from various smartphones (Samsung, IPhone, and Google Pixel), comprising a total of 11,016 images. This mix results in images of varying quality, which we organize into high-quality (smartphone) and low-quality (action camera) subsets. Of these 11,016 images, 10,004 correspond to nighttime scenes, including 6,004 images containing traffic signboards and 4,000 nighttime images without signboards. 

Furthermore, we provide auxiliary annotations along three axes: (i) image quality (smartphone vs.\ action-camera), (ii) weather condition (clear vs.\ rain), and (iii) per-image degradation attributes such as low illumination, blur, and glare. These subsets facilitate image-to-image translation tasks aimed at enhancing the visual quality of traffic or driving scenes. Besides nighttime images, we provide 1,012 day images captured from the same routes in both low-quality (from action camera) and high-quality (from smartphones) formats. These images can be utilized for day-to-night or night-to-day image translation tasks. We do not include a large set of daytime images due to their availability in prior Indian dataset \cite{uikey2024indian}.

\subsection{Traffic Sign Class Taxonomy}
We follow the official traffic sign taxonomy defined by the Ministry of Road Transport and Highways (MoRTH), Government of India \cite{morth2019roadsafety}, to map signboards to 41 class labels in INTSD. This set is selected in way that ensures there is no visual overlap within classes. It is important to highlight that the proposed taxonomy is inherently flexible, allowing for the incremental expansion of INTSD.

 \begin{figure*}[t]
    \centering
    \includegraphics[width=0.9\textwidth]{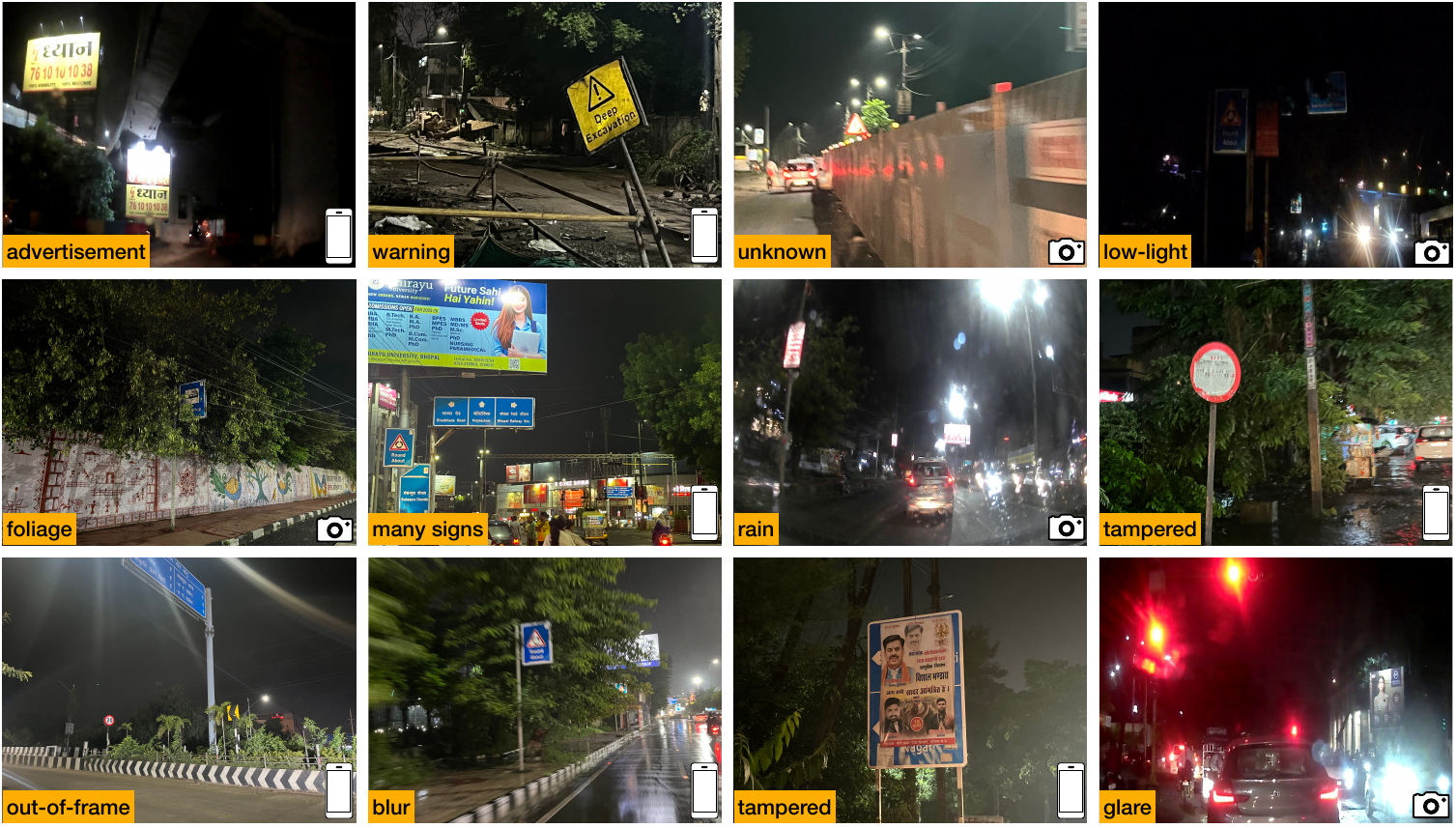}
    \caption{Examples from the INTSD. Each image highlights a representative attribute or challenge present in the dataset (e.g., glare, blur, rain, tampered signs, etc.). The icon in the bottom-right corner indicates the image source (action camera or smartphone), while the label denotes the corresponding visual attribute.}
    \label{fig:data-example}
\end{figure*}

\subsection{Annotation Process}
Annotating traffic signboards is a time-consuming process as it involves selecting and approving images, marking traffic signs with bounding boxes, and assigning appropriate class labels to each bounding box. To increase efficiency, we divide this process into three sequential stages: (i) image selection, (ii) manual annotation, and (iii) quality checks.

\noindent \textbf{Image Selection.} To manage the volume of raw footage, we first subsample at one frame per second, as adjacent frames at higher rates are visually near-identical and contribute redundant training signal. A filtering step was then applied to identify candidate frames containing traffic signs, followed by verification of every retained frame. Frames exhibiting severe glare or extreme motion blur (where no sign remained discernible) were discarded at this stage. However, INTSD intentionally retains images with light-to-moderate blur or glare when traffic signs remain visible, as such conditions frequently occur in real-world nighttime driving and introduce additional challenges for TSR.


\noindent \textbf{Manual Annotation.} We use the VGG Image Annotator (VIA) \cite{dutta2019via} to manually draw bounding boxes around traffic signboards and assign class labels to each corresponding box. All bounding boxes are subsequently verified for accuracy and completeness. INTSD includes bounding boxes for diverse traffic sign types, including directional, mandatory, informative, rural signs, highways signs, etc.

\noindent \textbf{Distractor Classes.}  In addition to the standard traffic sign classes, we introduce four supplementary classes to address cases that are commonly confused by the TSR models and are not represented in existing datasets: (i) the \textit{tampered} class includes signboards that are broken, rusted, vandalized, or covered with pamphlets; (ii) the \textit{unknown} class comprises of signboards that cannot be mapped to any existing class due to occlusion, motion blur, glare, small size, or partial visibility within the frame; (iii) the \textit{advertisement} class encompasses billboards and other large advertisement boards absent from current datasets; and (iv) the \textit{warning} class contains signboards indicating diversions or other road warnings. INTSD also includes images affected by moderate motion blur and light glare, reflecting real-world conditions and enhancing the model’s robustness to such visual distortions. We provide the visual examples of the distractor classes and challenges present in the INTSD in Figure \ref{fig:data-example}. Refer to the Section 1 of the supplementary material for more details and visuals.

\noindent \textbf{Quality Checks.} The annotations in INTSD follow a strict and standardized set of protocols to ensure high-quality labeling. Both detection and classification annotations are triple-verified to minimize the possibility of errors. Detection errors were minimal: 3.12\% false negatives were identified and added. During the second verification of class labels, 0.47\% labeling error rate was observed, which dropped to 0\% after the third (final) verification stage.




 \begin{figure*}[t]
    \centering
    \includegraphics[width=\textwidth]{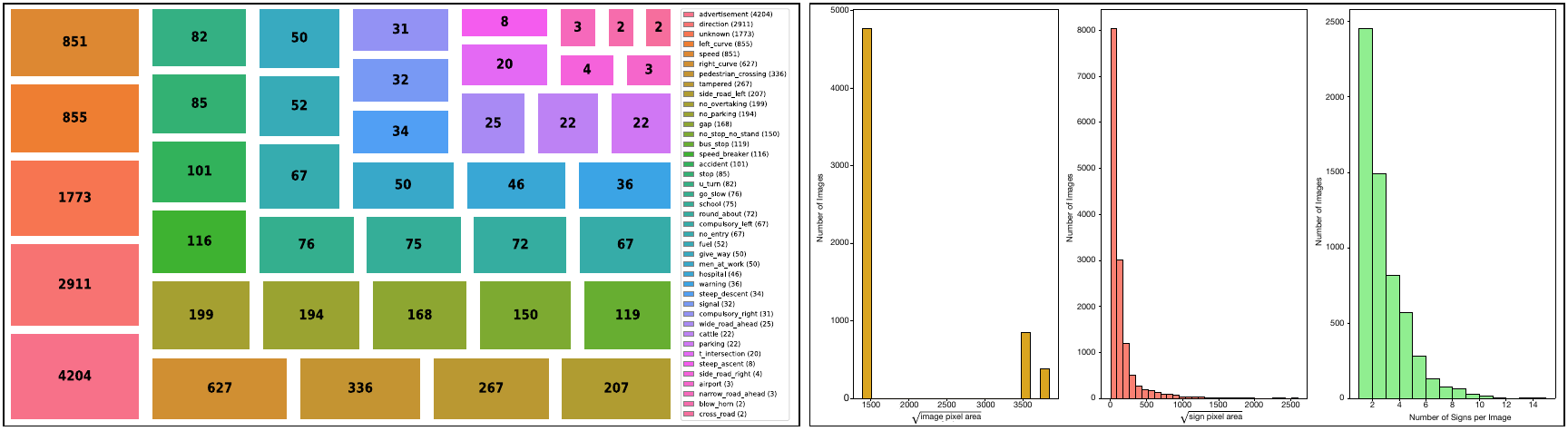}
    \caption{The tree map (on left) illustrates the class distribution. The bar graphs (on right) gives distribution of image size (golden) and signboard size (red) over $\sqrt{\text{pixel area}}$, and number of signboards per image (green).}
    \label{fig:data-stats}
\end{figure*}

\subsection{Data Splits}
We partition the dataset into five distinct splits. The five folds contain 1188, 1226, 1175, 1189, and 1226 images with 2133, 2134, 2129, 2133, and 2128 annotated signboard instances, respectively. Stratification preserves the class distribution in each fold, preventing rare classes from being underrepresented and maintaining a consistent distribution. The same holds true for the distractor classes. Fold definitions and annotation are publicly released to ensure reproducibility and facilitate standardized benchmarking.


\subsection{Statistics}
\label{subsec:stats}
The INTSD exhibits several characteristics that reflect the complexity of real-world nighttime sign recognition. We analyze key dataset statistics, including sensor diversity, class distribution, image resolution, signboard scale, object density, and geographical coverage. These statistics highlight the challenges posed by the dataset and motivate the need for robust detection and recognition methods.


\noindent \textbf{Sensor diversity.} INTSD is captured using a mixture of action cameras and smartphones, covering six different sensor types. This contrasts with datasets such as TT100K \cite{zhu2016traffic}, which rely on images captured from a single sensor setup. The resulting sensor diversity introduces variations in image quality, color response, and noise characteristics, contributing to a more diverse and realistic evaluation benchmark.

\noindent \textbf{Class Distribution.} Figure \ref{fig:data-stats} (left) provides the class distribution of INTSD. A primary characteristic of the dataset is a significant class imbalance, following a classic long-tail distribution. A small number of classes account for a large portion of the data. For instance, the top five most frequent classes: \textit{advertisement} (4,204 instances), \textit{direction} (2,911), \textit{unknown} (1,773), \textit{left\_curve} (855), and \textit{speed} (851), constitute a substantial majority of all labeled signs. Conversely, rare but crucial signs such as \textit{airport}, \textit{blow\_horn}, and \textit{cross\_road} are severely underrepresented. This imbalance poses a significant challenge, as standard training methods can become biased towards the majority classes, leading to poor performance on the rare but equally important minority classes. This characteristic strongly motivates the need for specialized methods that can learn robust representations for both the head and tail of the distribution.

\noindent \textbf{Image Sizes.} Figure \ref{fig:data-stats} (golden/right) depicts the bimodal distribution of the image sizes in INTSD. It emphasizes that the images primarily come in two distinct size ranges: a large number of smaller images (around $\sqrt{\text{area}} \approx 1100$) and another group of much larger images (around $\sqrt{\text{area}} \approx 3700$). The resulting variation in native resolution (differing levels of detail for small and distant signboards) makes INTSD a more realistic testbed for models that must operate robustly across heterogeneous capture devices.

\noindent\textbf{Signboard Sizes.} Figure \ref{fig:data-stats} (red/right) illustrates the long-tail distribution of signboard sizes in INTSD. As evident, the majority of signboards are relatively small, with a progressively smaller proportion of larger signboards. This inherent size imbalance poses a significant challenge for detection models, as small signboards contain limited visual information and are more susceptible to being overlooked by standard detection architectures. Such a distribution highlights the importance of employing strategies like scale-aware feature extraction and data augmentation to mitigate the bias towards more prevalent size categories.

\noindent \textbf{Signboards per Image.} Figure \ref{fig:data-stats} (green/right) also depicts a long-tail distribution of object density across the dataset with average of 2.32 signboards per image ($14044/6004$). Most images contain only a few signboards (1-4), whereas images with a higher number of signboards, corresponding to more cluttered scenes, are relatively less. This imbalance presents additional challenges for detection models, as densely populated scenes can introduce occlusions and complex spatial arrangements, making accurate localization and classification more difficult. Addressing this variation in object density requires models capable of robustly handling both sparse and cluttered scenarios.

\begin{figure*}[t]
    \centering
    \includegraphics[width=0.9\textwidth]{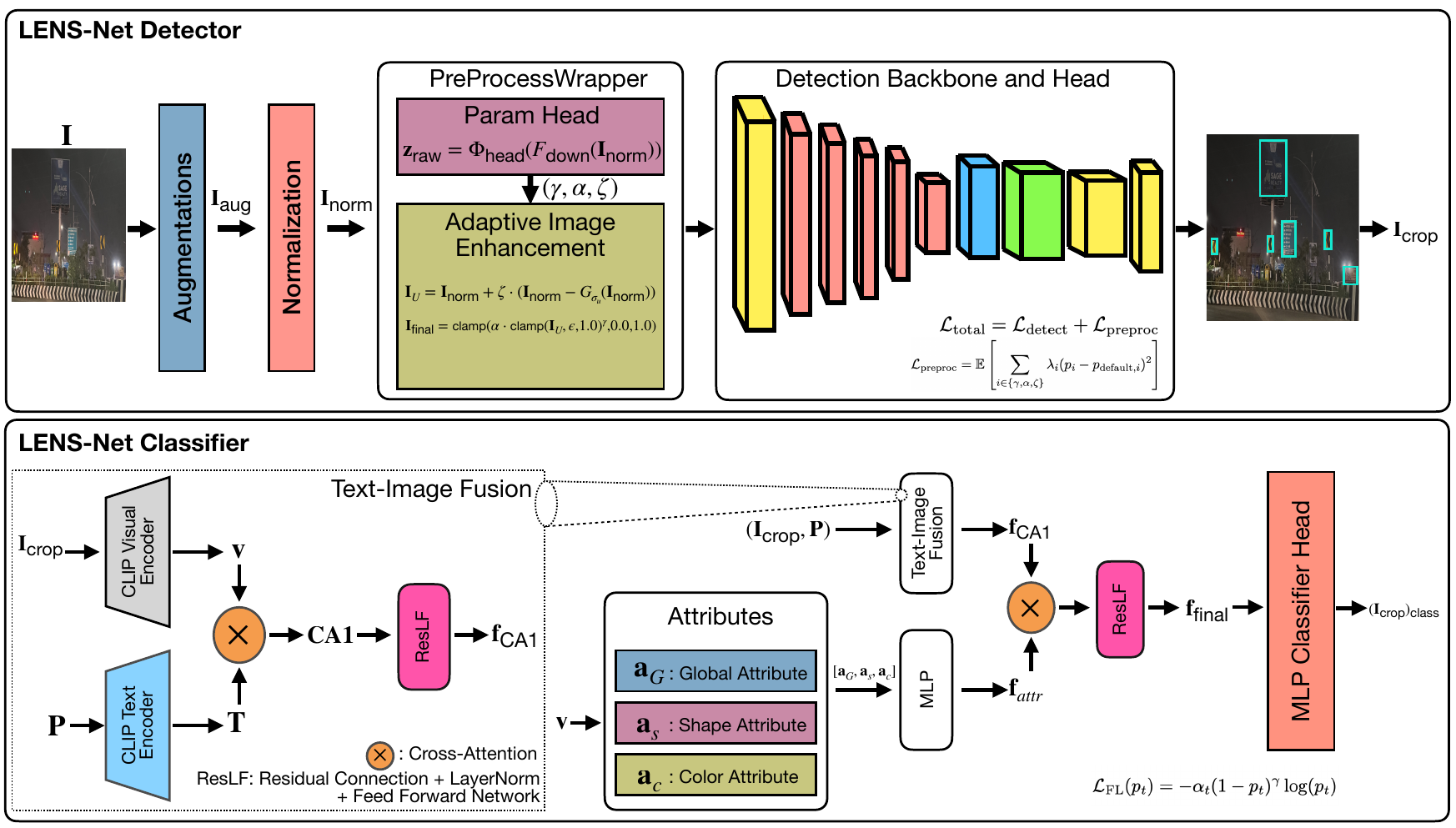}
    \caption{Proposed LENS-Net architecture: detector (top) integrates adaptive image enhancement for joint illumination correction and sign localization, and classifier (bottom) fuses vision-language representations with soft semantic attribute reasoning for sign classification.}
    \label{fig:lens-net}
\end{figure*}

\noindent \textbf{Geographical Distribution.} Figure \ref{fig:data-1.pdf} illustrates the geographical distribution of the images in INTSD. The images span six districts and encompass both rural and urban settings, as well as a diverse range of lighting conditions, including well-lit and poorly-lit environments. This diversity ensures that the dataset captures real-world variability.

Overall, these statistics demonstrate that INTSD captures diverse real-world conditions, making it a challenging benchmark for nighttime traffic sign recognition. 

\section{Method}
\label{sec:method}
We provide LENS-Net as an illumination-aware reference baseline for INTSD, not as the paper's central contribution. It follows a decoupled detect-then-classify design: an adaptive detector $\mathcal{D}$ localizes signboards under varying illumination, and a multimodal classifier $\mathcal{C}$ recognizes each cropped region. This separation lets each stage be optimized for its own objective. Figure~\ref{fig:lens-net} gives an overview; full implementation details are in the supplementary material.

\subsection{LENS-Net Detector}
We adopt YOLOv8 as the detection backbone for its strong performance on small, densely packed objects. Because INTSD spans a wide illumination range---from headlight- and street-lamp-lit scenes to severely underexposed ones---a static preprocessing pipeline is inadequate. We therefore wrap the backbone with a lightweight \textit{PreprocessWrapper}: a small parameter head $\Phi_{\text{head}}$ ($\approx$48.93k parameters) predicts per-image enhancement parameters (gamma correction $\gamma$, brightness $\alpha$, unsharp amount $\zeta$), which are applied through differentiable transforms before detection. Unlike standalone enhancement networks, $\Phi_{\text{head}}$ is trained end-to-end with the detection loss and adds negligible parameters.

To ensure that preprocessing module learns meaningful transformations and avoids collapsing to trivial solutions, we employ two key strategies. (i) \textit{Stochastic Augmentation:} During training, we apply a diverse set of strong, non-differentiable augmentations (photometric jitter, JPEG compression, Gaussian noise, blur, and small rotations) before the image is passed to the \textit{PreprocessWrapper}. It forces $\Phi_{\text{head}}$ to learn to invert these degradations and normalize the image to a consistent domain for the detector. (ii) \textit{Preprocessing Loss:} We apply an L2 regularization based loss ($\mathcal{L}_{\text{preproc}}$) that penalizes the predicted scaled parameters for deviating from their (no operation) default values: $\mathcal{L}_{\text{preproc}} \leftarrow \mathbb{E} \left[ \sum_{i \in \{\gamma, \alpha, \zeta\}} \lambda_i (p_i - p_{\text{default}, i})^2 \right]$, where $\lambda_i$ ($\lambda_i \in (0,0.01)$) are per-parameter loss weights. This loss is added to the standard YOLOv8 detection loss, $\mathcal{L}_{\text{detect}}$, during training. Therefore, the net loss for LENS-Net detector is: $\mathcal{L}_{\text{total}} \leftarrow \mathcal{L}_{\text{detect}} + \mathcal{L}_{\text{preproc}}$.

LENS-Net detector is trained using a zero-one scheduler: (i) head-only optimization with frozen backbone, and (ii) joint end-to-end training. The detector’s bounding-box output ($\mathbf{I}_{\text{crop}}$) is forwarded to the classifier for recognition.

\subsection{LENS-Net Classifier}
Fine-grained TSR under nighttime conditions demands more than visual appearance alone: signs sharing similar shapes (e.g., pedestrian crossing and school ahead) or colors (e.g., red-bordered warnings) are easily confused under low-light degradation. We therefore design a classifier that explicitly reasons over three complementary cues for each crop $\mathbf{I}_{\text{crop}}$: (i) a global semantic embedding from a frozen CLIP vision encoder, (ii) prompt-conditioned textual priors from the frozen CLIP text encoder, and (iii) soft shape/color attribute features. The two branches are fused by cross-attention (full procedure in the supplementary material).

\noindent\textbf{Branch 1: Text--image fusion.}  We process $\mathbf{I}_{\text{crop}}$ to obtain the corresponding global image embedding $\mathbf{v}$, which is fused with semantic context derived from a set of pre-defined textual prompts $\mathbf{P}$ (encoded into embeddings $\mathbf{T} \in \mathbb{R}^{n(\mathbf{P}) \times D}$ using the frozen CLIP text encoder) via cross-attention, followed by a combination of residual connection, layer normalization, and feedforward network block, yielding a text-conditioned feature $\mathbf{f}_{\text{CA1}}$ (Equation~\ref{eq:ca1}--\ref{eq:f_ca1}). This aligns the visual representation with textual cues useful for recognition (presence of numbers or arrows, etc.).

\begin{equation}
\label{eq:ca1}
    \mathbf{CA1} = \text{MHA}_{\text{CA1}}(\mathbf{Q}=\mathbf{v},\, \mathbf{K}=\mathbf{T},\, \mathbf{V}=\mathbf{T})
\end{equation}
\begin{equation}
\label{eq:f_ca1}
    \mathbf{f}_{\text{CA1}} = \text{FFN}(\text{LN}(\mathbf{v} + \mathbf{CA1})), \quad \mathbf{f}_{\text{CA1}} \in \mathbb{R}^{D}
\end{equation}

\noindent\textbf{Branch 2: Attribute module.} We explicitly model the two most discriminative sign attributes, shape and color. From $\mathbf{v}$, we predict soft distributions over learnable shape ($\mathbf{E}_S$) and color ($\mathbf{E}_C$) embedding tables, producing continuous attribute vectors $\mathbf{a}_S\; \text{and}\; \mathbf{a}_C$. Instead of making a hard, non-differentiable \textit{argmax} decision (e.g., this is 100\% a circle), we represent uncertainty (e.g., [0.85 (circle), 0.1 (octagon), 0.05 (other)]). Together with the global feature $\mathbf{a}_G = \mathbf{v}$, the three attribute vectors are concatenated and mixed by a 3-layer MLP, yielding refined attribute features $\mathbf{f}_{\text{attr}}$. (We found this simple MLP mixer matches a graph-based alternative while being simpler; see supplementary material).

\noindent\textbf{Fusion and classification.} The text-conditioned feature $\mathbf{f}_{\text{CA1}}$ attends to the attribute features $\mathbf{f}_{\text{attr}}$ via a second cross-attention block (Equation~\ref{eq:ca2}), and a residual + LayerNorm + FFN produces the final feature $\mathbf{f}_{\text{final}}$ (Equation~\ref{eq:f_final}), which an MLP head maps to class logits $(\mathbf{I}_{crop})_{\text{class}}$.

\begin{equation}
\label{eq:ca2}
    \mathbf{CA2} = \text{MHA}_{\text{CA2}}(\mathbf{Q}=\mathbf{f}_{\text{CA1}},\, \mathbf{K}=\mathbf{f}_{\text{attr}},\, \mathbf{V}=\mathbf{f}_{\text{attr}})
\end{equation}
\begin{equation}
\label{eq:f_final}
    \mathbf{f}_{\text{final}} = \text{FFN}(\text{LN}(\mathbf{f}_{\text{CA1}} + \mathbf{CA2}))
\end{equation}

\noindent\textbf{Training objective.} The classifier uses class-balanced focal loss~\cite{lin2017focal}, $\mathcal{L}_{\text{FL}}(p_t) = -\alpha_t (1-p_t)^\gamma \log(p_t)$, with weights $\alpha_t$ proportional to inverse square-root class frequency, plus a rarity-based augmentation. Here $p_t$ is the model's predicted probability for the ground-truth class, and $\gamma$ is the focusing parameter. Please refer to the  Section 2 of the supplementary material for additional details.

\section{Results}
\label{sec:results}

 \noindent \textbf{Experimental Settings.} We adopt a stratified 5-fold cross-validation strategy for training and evaluation, thereby reducing bias in the evaluation process. We choose mAP@50 and mAP@50:95 as detection evaluation metrics and macro-averaged precision (A.P.), macro-averaged recall (A.R.), and accuracy (Acc.) as classification evaluation metrics. We also report frames per second (FPS) to evaluate real-time performance and deployment feasibility. For fair comparison, FPS is measured end-to-end, from raw image loading to final predictions, including preprocessing, forward pass, post-processing, and CUDA synchronization. Each FPS value is averaged over five independent runs.

\noindent \textbf{Baseline Results.} 
In Tables~\ref{table:detect}--\ref{table:classify}, we benchmark INTSD using strong baselines covering multiple paradigms. For detection, we evaluate: (i) two-stage detectors (Faster R-CNN~\cite{ren2015faster}), (ii) one-stage detectors (YOLOv8~\cite{yolov8_ultralytics}, YOLOv10-L~\cite{wang2024yolov10}, YOLOv12-L~\cite{tian2025yolov12}, GOLD-YOLO-L~\cite{wang2023gold}), (iii) low-light–oriented detection methods (DAI-Net~\cite{du2024boosting}, YOLO-LLTS~\cite{lin2025yolo}, YOLO-TS~\cite{chen2025yolo}), and (iv) TSR approaches (YOLO-LLTS~\cite{lin2025yolo}, YOLO-TS~\cite{chen2025yolo}).

For classification, we compare against: (i) convolutional and transformer-based models (EfficientNet-B0~\cite{tan2019efficientnet}, DeIT~\cite{touvron2021training}), (ii) vision-language models (CLIP~\cite{radford2021learning}, BLIP~\cite{li2022blip}), (iii) YOLO based methods (YOLOv10-L~\cite{wang2024yolov10}, YOLOv12-L~\cite{tian2025yolov12}) and (iv) task-specific TSR approaches (Rishabh et al.~\cite{uikey2024indian}, YOLO-TS~\cite{chen2025yolo}). All models are trained and evaluated under identical data splits and protocols.

\begin{table}[t]
    \centering
    \footnotesize
    \caption{Detection performance (\%) on the INTSD.} 
    \label{table:detect}
    \setlength{\tabcolsep}{4pt}
    \resizebox{\columnwidth}{!}{
    \begin{tabular}{p{2.5cm}cccccc}
        \toprule
        \textbf{Model} & \textbf{mAP@50} & \textbf{mAP@50:95} & \textbf{FPS} & \multicolumn{2}{c}{\textbf{Parameters (M)}} \\
        \cmidrule(lr){5-6}
        & & & & \textbf{Total} & \textbf{Trainable} \\
        \midrule
        Faster R-CNN \cite{ren2015faster} & 62.60 & 43.22 & 23.38 & 41.50 & 41.50 \\
        DAI-Net \cite{du2024boosting} & 87.18 & 62.56 & 20.15 & 49.83 & 49.83 \\
        YOLOv12-L \cite{tian2025yolov12} & 82.84 & 64.93 & 59.08 & 26.48 & 26.48 \\
        GOLD-YOLO-L \cite{wang2023gold} & 88.70 & 72.11 & 42.64 & 74.99 & 74.99 \\
        YOLO-LLTS \cite{lin2025yolo} & 90.00 & 73.65 & 56.34 & 9.87 & 9.87 \\
        YOLOv10-L \cite{wang2024yolov10} & 90.00 & 75.81 & 64.39 & 25.83 & 25.83 \\
        YOLOv8 \cite{yolov8_ultralytics} & 90.13 & 77.21 & \textbf{83.71} & 11.14 & 11.14 \\
        YOLO-TS \cite{chen2025yolo} & 90.82 & 77.27 & 62.52 & 23.39 & 23.39 \\
        \midrule
        \textbf{LENS-Net} & \textbf{94.46} & \textbf{81.56} & 78.11 & 11.18 & 11.18 \\
        \bottomrule
    \end{tabular}
    }
\end{table}

\begin{table}[t]
    \centering
    \footnotesize
    \caption{Classification results (\%) on the INTSD.} 
    \label{table:classify}
    \setlength{\tabcolsep}{4pt}
    \resizebox{\columnwidth}{!}{
    \begin{tabular}{p{2.5cm}ccccccc}
        \toprule
        \textbf{Model} & \textbf{A.P.} & \textbf{A.R.} & \textbf{Acc.} & \textbf{FPS} & \multicolumn{2}{c}{\textbf{Parameters (M)}} \\
        \cmidrule(lr){6-7}
        & & & & & \textbf{Total} & \textbf{Trainable} \\
        \midrule
        YOLOv10-L \cite{wang2024yolov10} & 37.24 & 26.51 & 70.32 & 64.37 & 25.83 & 25.83 \\
        YOLOv12-L \cite{tian2025yolov12} & 45.70 & 39.60 & 67.66  & 59.14 & 26.48 & 26.48 \\
        BLIP \cite{li2022blip} & 49.33 & 47.81 & 69.90 & 75.37 & 4060.87 & 8.93 \\
        CLIP \cite{radford2021learning} & 65.25 & 68.11 & 82.49 & 72.37 & 434.04 & 12.85 \\
        DeIT \cite{touvron2021training} & 65.89 & 59.69  & 82.13 & 136.98 & 85.83 & 85.83 \\
        Rishabh et al. \cite{uikey2024indian} & 74.49 & 69.98 & 88.21 & 60.25 & 26.87 & 26.85 \\
        YOLO-TS \cite{chen2025yolo} & 75.08 & 74.43 & \textbf{89.70} & 63.36 & 23.39 & 23.39 \\
        EfficientNet-B0 \cite{tan2019efficientnet} & 81.68 & 75.42 & 89.12 & \textbf{188.44} & 4.06 & 4.06 \\ 
        \midrule
        \textbf{LENS-Net} & \textbf{85.09} & \textbf{79.32} & 88.38 & 64.33 & 453.54 & 25.93 \\
        \bottomrule
    \end{tabular}
    }
\end{table}

\noindent \textbf{Detection.} Table~\ref{table:detect} reports detection performance on INTSD across two-stage, one-stage, and low-light–oriented detectors. Modern YOLO variants achieve strong results, reflecting the dataset’s suitability for contemporary detection architectures. LENS-Net attains the highest mAP@50 (94.46\%) and mAP@50:95 (81.56\%), while maintaining competitive inference speed. The consistent performance gap over generic and low-light–specific detectors highlights the challenges of nighttime conditions and validates the benchmark's discriminative power. 

\noindent \textbf{Classification.}
Table~\ref{table:classify} presents classification performance across convolutional, transformer-based, vision-language, and task-specific TSR models. Vision-language approaches such as CLIP and BLIP provide strong semantic baselines, while task-specific models (e.g., Rishabh et al.~\cite{uikey2024indian}, YOLO-TS~\cite{chen2025yolo}) achieve higher task-adapted accuracy. LENS-Net achieves the best macro-averaged precision (85.09\%) and recall (79.32\%), demonstrating the effectiveness of structured multimodal reasoning under long-tailed and low-illumination conditions. Ablation study (see supplementary material) confirms the contribution of key components in LENS-Net classifier. These results establish competitive reference baselines for future research on INTSD. 

\begin{table}[t]
    \centering
    \caption{Cross-domain utility of INTSD on the 8-class shared taxonomy (classification on ground-truth crops). For each model we vary only the training data: \textit{Day} (Rishabh~et~al.\ daytime source), \textit{Joint} (day + a size-matched INTSD subset), and \textit{INTSD} (nighttime only). Values are mean\,$\pm$\,std over 5 folds. Legend: \cbox{softred} Day training, \cbox{softyellow} Joint training, \cbox{softgreen} nighttime training.}
    \label{tab:utility}
    \setlength{\tabcolsep}{4pt}
    \resizebox{\columnwidth}{!}{
    \begin{tabular}{lccccc}
        \toprule
        \textbf{Model} & \textbf{mAP@50} & \textbf{mAP@50:95} & \textbf{A.P.} & \textbf{A.R.} & \textbf{Acc.} \\
        \midrule
        \multirow{3}{*}{Rishabh et al.~\cite{uikey2024indian}}
          & \cellcolor{softred} -- & \cellcolor{softred} -- & \cellcolor{softred}61.03\, $\pm$1.33 & \cellcolor{softred}61.53\,$\pm$2.17 & \cellcolor{softred}71.63\, $\pm$0.77 \\
          
          &\cellcolor{softyellow} -- &\cellcolor{softyellow} --  & \cellcolor{softyellow}87.63\,$\pm$3.00 & \cellcolor{softyellow}92.70\, $\pm$0.88 & \cellcolor{softyellow}92.50\, $\pm$1.30 \\
          
          &\cellcolor{softgreen} -- & \cellcolor{softgreen} -- & \cellcolor{softgreen}94.87\, $\pm$2.96 & \cellcolor{softgreen}91.92\, $\pm$0.96  & \cellcolor{softgreen}96.41\, $\pm$0.93 \\
        \midrule
        \multirow{3}{*}{YOLO-TS~\cite{chen2025yolo}}
          &\cellcolor{softred}  5.31\, $\pm$1.02 &\cellcolor{softred} 1.59\, $\pm$0.37 &\cellcolor{softred} 54.01\, $\pm$4.9 &\cellcolor{softred} 58.48\,$\pm$2.76 &\cellcolor{softred} 71.81\,$\pm$0.92 \\
          
          &\cellcolor{softyellow} 91.82\,$\pm$1.04 &\cellcolor{softyellow} 76.07\,$\pm$1.57 &\cellcolor{softyellow} 89.38\,$\pm$1.70 &\cellcolor{softyellow} 91.80\,$\pm$1.45&\cellcolor{softyellow} 93.92\,$\pm$0.82 \\
          
          &\cellcolor{softgreen} 92.06\,$\pm$0.88 &\cellcolor{softgreen} 77.70\,$\pm$3.67 &\cellcolor{softgreen} 94.58\,$\pm$1.32 &\cellcolor{softgreen} 93.21\, $\pm$1.08 &\cellcolor{softgreen} 96.85\,$\pm$1.06 \\
        \midrule
        \multirow{3}{*}{LENS-Net}
          &\cellcolor{softred} 5.87\,$\pm$1.22 &\cellcolor{softred} 2.55\,$\pm$0.58 &\cellcolor{softred} 71.43\, $\pm$1.40 &\cellcolor{softred} 75.64\, $\pm$2.85 &\cellcolor{softred} 83.66\, $\pm$0.77 \\
          
          &\cellcolor{softyellow} 94.32\,$\pm$0.61 &\cellcolor{softyellow} 82.30\,$\pm$0.61 &\cellcolor{softyellow} 80.44\, $\pm$1.61  &\cellcolor{softyellow} 86.71\, $\pm$1.64 &\cellcolor{softyellow} 88.98\, $\pm$2.18 \\
          
          &\cellcolor{softgreen} 93.82\,$\pm$0.91 &\cellcolor{softgreen} 81.27\,$\pm$1.93 &\cellcolor{softgreen} 91.89\, $\pm$1.50 &\cellcolor{softgreen} 89.22\, $\pm$1.25 &\cellcolor{softgreen} 92.95\,$\pm$1.99  \\
        \bottomrule
    \end{tabular}
    }
\end{table}

\begin{table}[t]
\centering
\caption{Detection and classification performance on CNTSSS. $ ^{\dag}$ YOLO-LLTS is detection-only model.}
\Large
\label{tab:cntsss}
\setlength{\tabcolsep}{4pt}
\resizebox{\columnwidth}{!}{
\begin{tabular}{lccccccc}
\toprule
\multirow{2}{*}{\textbf{Model}} & 
\multicolumn{3}{c}{\textbf{Detection (\%)}} & 
\multicolumn{4}{c}{\textbf{Classification (\%)}} \\

 & \textbf{mAP@50} & \textbf{mAP@50:95} & \textbf{FPS}
 & \textbf{A.P.} & \textbf{A.R.} & \textbf{Acc.} & \textbf{FPS} \\

\midrule
YOLOv10-L \cite{wang2024yolov10} & 68.84 & 46.56 & 100.51 & 84.19 & 84.15 & 86.02 & 100.32 \\
YOLO-LLTS$^{\dag}$ \cite{lin2025yolo} & 76.30 & 54.15 & 72.17 & -- & -- & -- & -- \\
YOLO-TS \cite{chen2025yolo}  & \textbf{76.47} & \textbf{54.21} & 102.51 & 83.92 & 84.08 & 86.46 & 102.54 \\
\midrule
\textbf{LENS-Net}  & 69.17 & 46.62 & \textbf{143.85} & \textbf{87.35} & \textbf{88.64} & \textbf{89.70} & 74.43 \\

\bottomrule
\end{tabular}
}
\end{table}

\noindent \textbf{Data Utility.} Table~\ref{tab:utility} evaluates cross-domain utility on an 8-class shared taxonomy (intersection of INTSD and the daytime data of Rishabh~et~al.~\cite{uikey2024indian}). The 8 shared classes are a strict subset, so any gain from introducing INTSD cannot be attributed to class coverage. Three training regimes are compared across all architectures: \textit{Day} (daytime data only), \textit{Joint} (day + a size-matched INTSD subset), and \textit{INTSD} (nighttime only), where the Joint condition controls for data volume so recovery cannot be credited to quantity alone. All the models are tested on the 8-class subset of INTSD across five folds. The pattern is consistent across all three models: day-trained models fail substantially at night — detection collapses and classification degrades for all architectures. The performance recovers once INTSD is introduced in either the Joint or INTSD-only regime. Notably, LENS-Net retains comparatively stronger classification performance even under the day-only regime (71.43\% A.P.) relative to the other models, reflecting its semantic attribute reasoning. YOLO-TS~\cite{chen2025yolo} and Rishabh~et~al.~\cite{uikey2024indian}\ achieve higher scores than LENS-Net on this 8-class subset when trained on nighttime data, however this advantage does not transfer to the full 41-class INTSD benchmark, as shown in Table~\ref{table:classify}, where LENS-Net outperforms both.

\noindent \textbf{Generalization to CNTSSS.}
Table~\ref{tab:cntsss} reports cross-dataset evaluation on CNTSSS~\cite{lin2025yolo}. While several methods achieve strong performance due to its limited class taxonomy, INTSD produces larger performance variation across models, reflecting increased semantic diversity and intra-class ambiguity. LENS-Net achieves competitive detection and the highest classification performance, demonstrating robustness across differing nighttime benchmarks.

To validate INTSD's utility as a complementary training resource, we additionally conduct joint training under a unified 44-class label space on CNTSSS~\cite{lin2025yolo} and INTSD (evaluated on held-out test sets from both; Section~3 of supplementary material). Results confirm that incorporating INTSD improves cross-domain robustness without degrading CNTSSS performance, supporting its role as a regularizing source of diverse nighttime conditions.


\section{Conclusion}
We present INTSD, the first large-scale Indian nighttime traffic sign dataset, comprising 11k+ images and 14k+ annotated instances across 41 classes under diverse illumination, weather, and sensor conditions. Through extensive benchmarking, we demonstrate that daytime-trained models fail substantially under real nighttime conditions, and that this gap is reliably recovered by incorporating INTSD — validating its complementary utility as a nighttime training resource. To accompany the benchmark, we provide LENS-Net, an illumination-aware multimodal baseline integrating adaptive enhancement with vision-language and semantic attribute reasoning, achieving the best performance among all evaluated models. Together, INTSD and LENS-Net form a standardized, reproducible evaluation platform for robust traffic sign recognition under real-world nighttime conditions that existing daytime benchmarks cannot capture.

{
    \small
    \bibliographystyle{ieeenat_fullname}
    \bibliography{main}
}

\end{document}